\documentclass[a4paper,fleqn]{cas-sc}

\usepackage[authoryear]{natbib}
\usepackage{caption}

\def\tsc#1{\csdef{#1}{\textsc{\lowercase{#1}}\xspace}}
\tsc{WGM}
\tsc{QE}
\tsc{EP}
\tsc{PMS}
\tsc{BEC}
\tsc{DE}

\begin{document}
\let\WriteBookmarks\relax
\def\floatpagepagefraction{1}
\def\textpagefraction{.001}
\shorttitle{G-SHARE for Human-Factor Diagnosis}
\shortauthors{Xiao et~al.}

\title{G-SHARE: A Guideline-Based Structured Reasoning Framework for Human-Factor Event Diagnosis}

\tnotetext[1]{The research was supported by a grant from  the National Natural Science Foundation of China (Grant No. T2192933), the LingChuang Research Project of China National Nuclear Corporation, the Foundation of National Key Laboratory of Human Factors Engineering (Grant No. HFNKL2024W07), and Tsinghua University Initiative Scientific Research Program.}

\author[1,2]{Xingyu Xiao}[style=chinese]

\credit{Conceptualization, Methodology, Software, Formal analysis, Data Curation, Visualization, Validation, Writing- Original draft preparation.}

\author[3]{Mao Du}[style=chinese]
\credit{Investigation, Resources, Validation, Writing - review and editing.}

\author[1,2]{Jiejuan Tong}[style=chinese]
\credit{Conceptualization, Formal analysis, Supervision, Writing - Review and Editing.}

\author[1]{Jingang Liang}[style=chinese]
\credit{Supervision, Writing- Reviewing and Editing.}

\author[1]{Haitao Wang}[style=chinese]
\credit{Supervision, Writing- Reviewing and Editing.}

\affiliation[1]{organization={Institute of Nuclear and New Energy Technology, Tsinghua University},
            city={Beijing},
            postcode={100084}, 
            country={China}}
\affiliation[2]{organization={National Key Laboratory of Human Factors Engineering},
            city={Beijing},
            postcode={100094}, 
            country={China}}

\affiliation[3]{organization={Fujian Fuqing Nuclear Power Co., Ltd.},
            city={Fujian},
            postcode={350318‌}, 
            country={China}}
            
\begin{abstract}
Human-factor event diagnosis is essential for learning from operational events in nuclear power plants, yet its quality depends strongly on expert interpretation of narrative reports and guideline-based reasoning.Existing data-driven or one-shot large language model approaches often lack structured reasoning, have limited alignment with formal diagnostic guidelines, and may generate logically inconsistent conclusions. To address this issue, this study proposes G-SHARE, a guideline-based structured reasoning framework that operationalizes the CNNP nine-step human-factor event diagnosis guideline into a multi-stage diagnostic pipeline.The framework consists of evidence extraction, stepwise diagnostic reasoning, and post-hoc consistency repair, enabling explicit use of report evidence, intermediate rationale generation, and logical validation of diagnostic outputs. A dataset of real human-factor event reports was constructed from Chinese nuclear industry sources, and a gold-standard subset annotated by domain experts was used for evaluation. Results show that G-SHARE substantially outperforms one-shot prompting and traditional machine learning baselines, with the strongest version achieving the best overall accuracy and macro-F1. Ablation results further indicate that structured reasoning and consistency enforcement are critical to robust diagnosis, especially under weak prompting conditions. The findings demonstrate the value of transforming expert diagnostic guidelines into auditable reasoning workflows, providing a practical pathway for intelligent human-factor analysis in safety-critical industries.

\end{abstract}


\maketitle

\section{Introduction}

Human-factor event diagnosis plays a critical role in learning from operational events and enhancing safety performance in nuclear power plants \cite{poller2020human}. Following abnormal events, systematic analysis is conducted to identify human errors, understand their underlying mechanisms, and support the development of corrective and preventive measures \cite{anu2018development}. In practice, this process extends beyond simple categorization, requiring analysts to reconstruct event sequences, interpret operator behaviors, and evaluate cognitive and situational factors in accordance with established diagnostic guidelines \cite{kim2019influencing}. However, event reports are typically narrative and semi-structured, with key information embedded in dispersed textual descriptions \cite{mueller2019episodic}. As a result, analysts must manually extract relevant evidence and synthesize it into a coherent diagnostic interpretation. Although formal procedures—such as the nine-step human-factor event diagnosis guideline developed in the Chinese nuclear industry—provide structured analytical principles, their execution remains highly dependent on expert judgment \cite{mayfield2008organizational}. This leads to limitations in efficiency, consistency, and reproducibility, especially when dealing with large volumes of event data \cite{zwanenburg2019radiomics}. Therefore, there is a need for a more structured and reproducible approach to human-factor event diagnosis that can preserve expert reasoning logic while reducing reliance on purely manual analysis \cite{zarei2023account}.

Recent advances in artificial intelligence have motivated efforts to support or automate event analysis \cite{holmes2004artificial}. Traditional machine learning approaches typically formulate the task as a text classification problem, directly mapping event reports to predefined diagnostic categories \cite{young2019systematic}. While such methods can achieve reasonable predictive performance, they lack explicit reasoning processes and do not incorporate domain-specific diagnostic guidelines, making their outputs difficult to interpret and verify \cite{arocha2005identifying}. More recently, large language models have been applied to event analysis using free-form prompting, leveraging their strong language understanding capabilities to generate diagnostic conclusions \cite{da2025flans}. However, such approaches often produce unconstrained outputs, which may overlook critical evidence, exhibit inconsistencies in reasoning, or fail to align conclusions with intermediate analysis \cite{hunter1998managing}. In safety-critical contexts, these limitations raise concerns regarding reliability, transparency, and auditability \cite{rausand2014reliability}. Meanwhile, existing human reliability analysis methods and event analysis tools provide structured frameworks for evaluating human performance, but are not designed to process narrative reports through guideline-based reasoning \cite{lekadir2025future}. In particular, there remains a lack of approaches that explicitly translate formal diagnostic guidelines into machine-executable reasoning processes. Consequently, the key limitation of current approaches is not merely predictive accuracy, but the absence of a structured and guideline-aligned reasoning mechanism that supports transparent and reliable human-factor event diagnosis \cite{poller2020human}.

In industrial practice, diagnostic guidelines such as the CNNP nine-step human-factor event diagnosis procedure represent a valuable accumulation of expert knowledge \cite{tang2018automatic}, defining how event information should be interpreted, how causal factors should be identified \cite{meyer2011experience}, and how human errors should be categorized. However, these guidelines are primarily intended for manual use and have not been fully operationalized into computational frameworks \cite{mitchell2014review}. Specifically, there is a lack of methods capable of transforming such guidelines into machine-readable reasoning stages, systematically extracting and utilizing evidence from narrative reports, enforcing consistency between intermediate reasoning and final conclusions \cite{mayer2021enhancing}, and producing reproducible and auditable diagnostic outputs. Addressing this gap requires bridging the divide between expert-defined analytical procedures and data-driven or language-based models \cite{withers2025natural}. A central question, therefore, is how domain diagnostic guidelines can be translated into structured reasoning frameworks that preserve expert logic while enabling scalable and auditable AI-assisted analysis \cite{surisetty2025ai}.

To address this challenge, this study proposes a guideline-based structured reasoning framework for human-factor event diagnosis, termed G-SHARE. The framework translates the CNNP nine-step diagnostic guideline into a multi-stage reasoning process, integrating evidence extraction, stepwise diagnosis, and logic-based consistency repair. In doing so, it enables explicit use of report evidence, structured generation of intermediate reasoning, and validation of diagnostic consistency. A dataset of real-world human-factor event reports is constructed, along with a gold-standard subset annotated by domain experts for evaluation. The results demonstrate that the proposed framework improves diagnostic performance compared with one-shot large language model approaches and traditional machine learning methods, while also providing more transparent and auditable reasoning processes. More broadly, this work illustrates a practical pathway for transforming expert diagnostic guidelines into machine-executable workflows, contributing to the development of trustworthy AI-assisted analysis in safety-critical domains. The remainder of this paper is organized as follows. Section 2 introduces the background and problem formulation. Section 3 presents the proposed framework. Section 4 describes the experimental design, followed by results in Section 5. Section 6 discusses the implications and limitations of this work, and Section 7 concludes the paper.

\section{Related Work}\label{sec:related}

\subsection{Human-factor event analysis and diagnosis methods}

Human-factor event analysis is a fundamental component of safety management in high-reliability industries such as nuclear power, aviation, and chemical processing \cite{bevilacqua2018human}. Over the past decades, a variety of human reliability analysis (HRA) methods have been developed to support the identification, quantification, and interpretation of human errors \cite{levine2024identifying}. Classical approaches, including THERP, SPAR-H, and more recent frameworks such as IDHEAS, provide structured procedures for analyzing human performance by decomposing tasks, identifying performance shaping factors \cite{pt2002therapeutic}, and estimating error probabilities. These methods have contributed significantly to improving the systematic understanding of human error mechanisms and their impact on system safety \cite{gertman1992intent}.

In addition to probabilistic HRA methods, event investigation and diagnosis frameworks are widely applied in operational practice to analyze real-world incidents \cite{ahmad2021can}. These approaches typically focus on reconstructing event sequences, identifying causal factors, and determining the nature of human errors through structured analytical procedures \cite{strauch2017investigating}. In the context of nuclear power, diagnostic guidelines—such as the nine-step human-factor event diagnosis procedure developed within the Chinese nuclear industry—provide detailed instructions on how analysts should extract evidence from event reports, interpret operator behaviors, assess contextual conditions \cite{stanton2017human}, and ultimately classify human-factor events. Such guidelines emphasize a stepwise reasoning process that integrates multiple sources of information and supports consistent decision-making \cite{hasic2018augmenting}.

Despite their strengths, existing human-factor analysis and diagnosis methods are predominantly designed for manual execution and rely heavily on expert judgment \cite{stanton2017human}. The analysis process is often time-consuming and requires substantial domain expertise, particularly when dealing with narrative and semi-structured event reports \cite{henriksen2022methods}. Moreover, variability across analysts may lead to inconsistencies in diagnostic outcomes, and the reasoning process is not always fully documented in a reproducible manner \cite{shinkins2013diagnostic}. While these approaches provide well-defined analytical principles, they lack computational mechanisms to directly operationalize diagnostic procedures into machine-executable workflows \cite{liu2018generating}.

Consequently, there remains a gap between formal diagnostic guidelines and their practical implementation in large-scale or data-driven analysis settings \cite{ikegwu2022big}. Existing methods offer structured conceptual frameworks, but do not provide explicit support for transforming guideline-based reasoning into automated, reproducible, and auditable diagnostic processes \cite{lichtner2023automated}.

\subsection{AI-based approaches for event analysis and text-based diagnosis}

With the increasing availability of operational data and advances in natural language processing, artificial intelligence (AI) techniques have been explored to support event analysis and human-factor diagnosis \cite{mah2022natural}. Early studies primarily adopted traditional machine learning approaches, in which textual event reports were transformed into feature representations, such as term frequency–inverse document frequency (TF-IDF) \cite{al2022document}, and subsequently classified using models including logistic regression, random forests, and gradient boosting methods \cite{lombardo2015binary}. These approaches enable automated processing of large volumes of reports and can achieve reasonable predictive performance. However, they typically treat the task as a direct mapping from text to labels \cite{minaee2021deep}, without explicitly modeling the underlying reasoning process or incorporating domain-specific diagnostic knowledge. As a result, their outputs are often difficult to interpret and may not align with established analytical procedures \cite{fortsch2018systematizing}.

More recently, large language models (LLMs) have demonstrated strong capabilities in understanding and generating natural language, leading to their application in event analysis tasks \cite{karanikolas2023large}. By leveraging prompt-based techniques, LLMs can produce diagnostic conclusions along with textual explanations, offering a more flexible alternative to traditional classifiers \cite{liu2019towards}. In particular, one-shot or few-shot prompting enables rapid deployment without extensive task-specific training, making these models attractive for practical applications \cite{hogstrom2010create}. Nevertheless, such approaches are typically based on free-form generation, where the reasoning process is not explicitly constrained. Consequently, the models may overlook critical evidence, produce incomplete or inconsistent reasoning chains, or generate conclusions that are not fully supported by the input reports \cite{sun2025survey}. These issues are particularly concerning in safety-critical contexts, where diagnostic results must be reliable, transparent, and subject to verification \cite{height2012practical}.

In parallel, AI techniques have also been applied to broader safety analysis tasks, such as accident report classification, risk factor identification \cite{wang2022artificial}, and event summarization. While these studies demonstrate the potential of AI for processing unstructured safety data, they generally focus on improving predictive accuracy or text understanding, rather than replicating the structured reasoning processes used by domain experts \cite{tenenbaum2011grow}. In particular, limited attention has been paid to integrating formal diagnostic guidelines into AI models \cite{sharma2025integration}, or to ensuring consistency between intermediate reasoning steps and final diagnostic outcomes.

Overall, existing AI-based approaches provide powerful tools for handling textual data, but they do not fully address the requirements of human-factor event diagnosis in safety-critical environments \cite{badhan2025artificial}. The key limitation lies not only in model performance, but in the absence of structured and guideline-aligned reasoning mechanisms that can support transparent, consistent, and auditable diagnostic processes \cite{lin2025performance}.

\subsection{Structured reasoning and guideline-driven AI systems}

To improve the reliability and interpretability of AI systems, recent studies have explored structured reasoning approaches that explicitly decompose complex tasks into intermediate steps \cite{li2022interpretable}. Instead of relying on direct input–output mappings, these methods guide models to generate stepwise reasoning processes, such as chain-of-thought prompting, program-of-thought reasoning, and tool-augmented inference \cite{zhang2025igniting}. By making intermediate reasoning explicit, structured approaches can enhance transparency and provide opportunities for verification, which is particularly important in tasks requiring logical consistency and multi-stage decision-making \cite{shiffman2012building}.

In parallel, there has been growing interest in integrating domain knowledge and constraints into AI systems \cite{nie2025data}. Approaches such as knowledge graph–enhanced models, rule-based reasoning, and constraint-guided inference aim to incorporate prior knowledge into the reasoning process \cite{ke2024enabling}, thereby improving robustness and alignment with domain-specific requirements. In safety-critical applications, additional mechanisms—such as verification modules and post-hoc consistency checking—have been proposed to ensure that model outputs satisfy predefined logical or operational constraints \cite{tambon2022certify}. These developments highlight the importance of combining data-driven models with structured knowledge representations to achieve more trustworthy and controllable AI behavior \cite{zha2025data}.

Despite these advances, existing structured reasoning and knowledge-guided approaches are largely designed as general-purpose techniques and are not specifically tailored to domain-defined diagnostic procedures \cite{cheng2025survey}. In particular, limited attention has been given to transforming formal guidelines into executable reasoning workflows that reflect how domain experts perform analysis in practice \cite{sethi2017scientific}. As a result, key aspects of human-factor event diagnosis—such as systematic evidence utilization, stepwise causal interpretation, and consistency between intermediate reasoning and final conclusions—are not fully captured by current methods \cite{mitchell2014review}.

Therefore, there remains a need for approaches that bridge structured reasoning techniques with domain-specific diagnostic guidelines, enabling the construction of reasoning frameworks that are both computationally tractable and aligned with expert analytical procedures \cite{gao2023dr}. Such integration is essential for supporting transparent, reproducible, and auditable decision-making in safety-critical human-factor analysis.

\section{Problem Formulation}\label{sec:problem}

\subsection{Human-factor event diagnosis in operational practice}

Human-factor event diagnosis in nuclear power plants is a structured analytical process aimed at identifying the nature and causes of human errors from operational event reports. In practical settings, the primary input to this process is a narrative event report, which documents the sequence of events, operator actions, system responses, and contextual conditions associated with an abnormal occurrence. These reports are typically written in semi-structured natural language and may vary in level of detail, clarity, and organization depending on the reporting context.

The objective of diagnosis is to determine the type of human-factor event or error mechanism associated with the reported incident. This involves not only assigning a diagnostic category, but also interpreting the underlying behavioral and cognitive processes that contributed to the event. In established industrial practice, such diagnosis is guided by formal procedures, which require analysts to examine evidence, reconstruct event progression, identify contributing factors, and evaluate the consistency between observed actions and expected operational behavior.

Despite the existence of standardized guidelines, human-factor event diagnosis remains a challenging task. First, the information contained in event reports is often distributed across multiple textual segments, with key evidence embedded in descriptive narratives rather than explicitly structured fields. Analysts must therefore identify and extract relevant information, such as operator intentions, system states, and temporal relationships, from potentially noisy or incomplete descriptions. Second, the diagnostic process requires integrating multiple types of evidence, including procedural deviations, contextual constraints, and system feedback, into a coherent causal interpretation. This integration is inherently complex and depends on both domain knowledge and reasoning ability.

Furthermore, the diagnosis procedure itself follows a stepwise logic defined by domain guidelines, which involves multiple intermediate judgments before reaching a final conclusion. These steps are not independent; rather, they are interrelated and must be internally consistent. In practice, ensuring such consistency is nontrivial, particularly when different pieces of evidence may suggest competing interpretations. As a result, diagnostic outcomes may vary across analysts, and the reasoning process is not always fully transparent or reproducible.

Taken together, human-factor event diagnosis can be viewed as a multi-stage reasoning task that requires structured evidence extraction, guideline-based interpretation, and consistency-aware decision-making. These characteristics highlight the need for approaches that can explicitly model and support the diagnostic process, rather than treating it as a direct text classification problem.

\subsection{Nine-step diagnostic guideline and its computational interpretation}

In industrial practice, human-factor event diagnosis is typically guided by formalized analytical procedures that define how analysts should systematically interpret event information and derive diagnostic conclusions. These procedures are often organized as multi-step guidelines, which decompose the diagnosis process into a sequence of structured reasoning tasks. Rather than directly assigning a diagnostic label, analysts are required to follow a stepwise workflow that progressively refines their understanding of the event.

The guideline considered in this study consists of nine sequential steps, covering the complete diagnostic process from initial evidence identification to final category determination. Although originally designed for manual analysis, the guideline embodies a well-defined logical structure that can be interpreted as a multi-stage reasoning process. Specifically, the nine steps can be grouped into several functional components.

\begin{center}
\includegraphics[width=1.0 \textwidth]{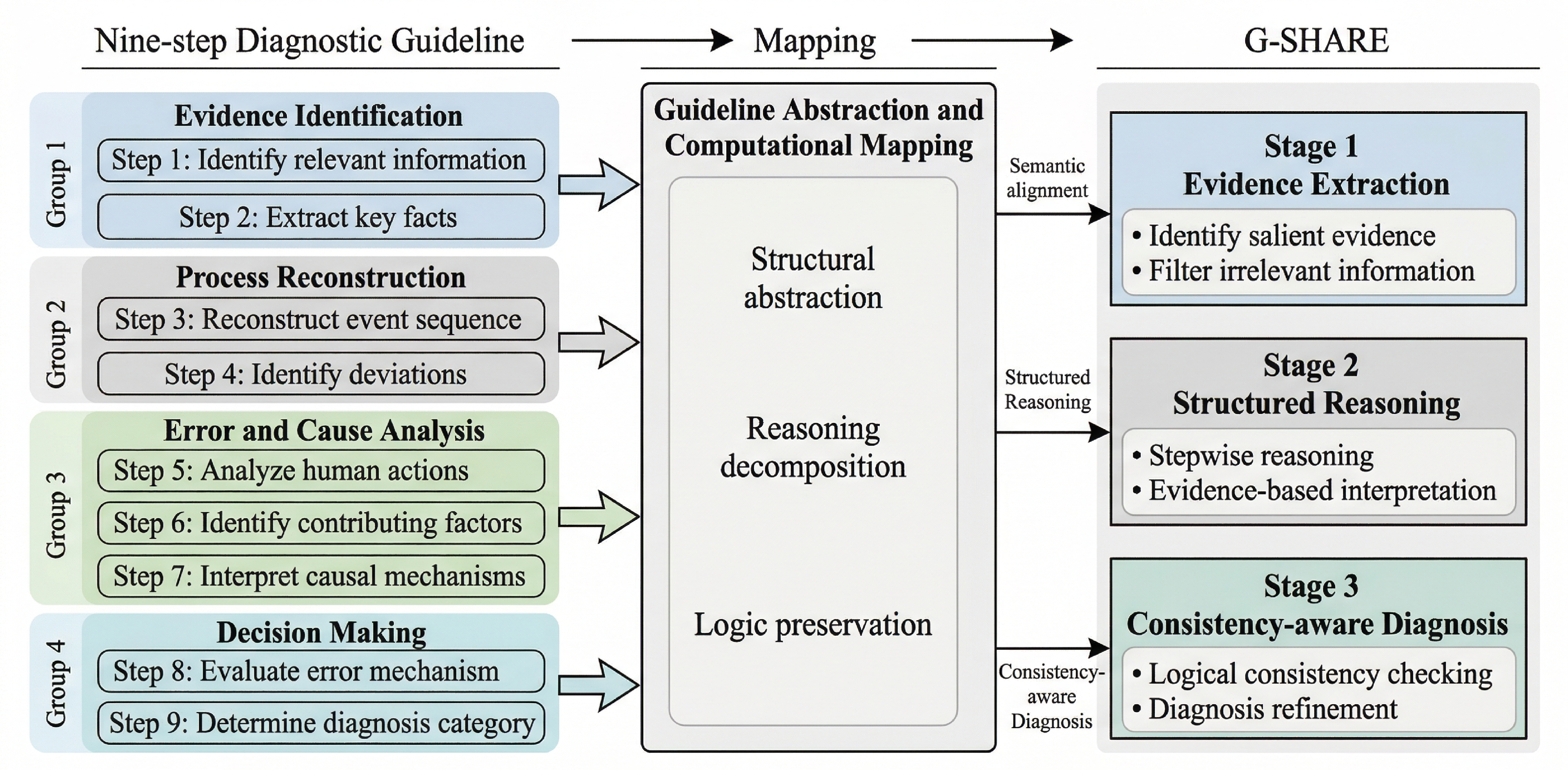}
\captionof{figure}{Transformation from a nine-step human-factor diagnostic guideline to structured computational reasoning stages in G-SHARE.}
\label{fig:framework_overview}
\end{center}

First, the initial steps focus on \emph{evidence identification}, where analysts extract relevant information from narrative event reports, including operator actions, system states, and contextual conditions. This stage aims to transform unstructured textual descriptions into a set of salient facts that are essential for subsequent analysis.

Second, the guideline emphasizes \emph{event process reconstruction}, in which the temporal sequence of events and interactions between operators and systems are organized into a coherent representation. This reconstruction enables analysts to understand how the event unfolded and to identify critical points where deviations occurred.

Third, the intermediate steps involve \emph{error attribution} and \emph{causal interpretation}. At this stage, analysts examine the reconstructed process to determine the nature of human errors, evaluate contributing factors, and interpret the relationship between observed actions and underlying cognitive or situational mechanisms. This requires integrating multiple sources of evidence and applying domain knowledge to distinguish between different types of human-factor events.

Finally, the guideline concludes with \emph{category judgment}, where the event is assigned to a predefined diagnostic category based on the preceding analysis. This decision must be consistent with the identified evidence and the inferred causal mechanisms, ensuring coherence between intermediate reasoning and final outcomes.

Although the nine-step guideline provides a comprehensive framework for expert analysis, it is primarily intended for manual execution and does not directly support computational implementation. However, its underlying structure reveals a natural decomposition into a limited number of reasoning stages, each corresponding to a distinct analytical function. This observation enables the transformation of the guideline into a computational framework, where the original procedural steps are mapped onto a sequence of machine-executable reasoning modules.

As illustrated in Fig.~\ref{fig:framework_overview}, the nine-step diagnostic guideline can be abstracted into three core stages: evidence extraction, structured diagnostic reasoning, and consistency-aware decision making. This abstraction preserves the logical flow of the original procedure while enabling its integration with AI-based methods, forming the basis for the proposed framework.

\subsection{Task formulation}

Based on the operational characteristics of human-factor event diagnosis and the structure of the underlying diagnostic procedure, the problem can be formulated as a structured reasoning task with explicit intermediate representations.

Let $x = (b, s)$ denote a human-factor event report, where $b$ represents the behavioral descriptor and $s$ denotes the narrative summary. The objective is to assign the report to a human-factor diagnosis label $y \in \mathcal{Y}$, where $\mathcal{Y} = \{y_1, \dots, y_7\}$ is the predefined set of root-cause categories.

Unlike conventional text classification problems, the mapping from $x$ to $y$ is not performed directly. Instead, the diagnosis process involves a sequence of intermediate reasoning steps that reflect the structure of the diagnostic guideline. To capture this process, we introduce two intermediate variables: an evidence set and a reasoning trace.

First, an \emph{evidence set} $E = \{e_1, e_2, \dots, e_m\}$ is extracted from the input report, where each element $e_i$ corresponds to a salient piece of information, such as operator actions, system states, or contextual conditions. The role of $E$ is to provide a structured representation of relevant information that supports subsequent reasoning.

Second, a \emph{reasoning trace} $R = (r_1, r_2, \dots, r_k)$ is constructed, where each step $r_i$ represents an intermediate judgment aligned with the diagnostic procedure. In particular, $R$ can be viewed as a sequence of answers to structured diagnostic questions, encoding the stepwise interpretation of the event. This representation makes the reasoning process explicit and enables verification of intermediate decisions.

Given these intermediate variables, the diagnostic process can be expressed as a two-stage mapping:
\begin{equation}
    (E, R) = f_{\text{reason}}(x), \quad y = f_{\text{dec}}(R),
\end{equation}
where $f_{\text{reason}}(\cdot)$ denotes the structured reasoning function that generates evidence and intermediate judgments, and $f_{\text{dec}}(\cdot)$ represents the decision function that maps the reasoning trace to a final diagnosis label.

However, due to the complexity of the reasoning process and the potential uncertainty in intermediate steps, the initial diagnosis $y$ may not always satisfy the logical constraints implied by the diagnostic procedure. To address this issue, we further introduce a consistency-aware refinement step, which produces a repaired output $\hat{y}$:
\begin{equation}
    \hat{y} = f_{\text{repair}}(y, R).
\end{equation}
Here, $f_{\text{repair}}(\cdot)$ enforces logical consistency between intermediate reasoning steps and the final diagnosis, ensuring that the output adheres to the constraints of the underlying decision structure.

Overall, the human-factor event diagnosis task can be formulated as a structured mapping:
\begin{equation}
    x \rightarrow E \rightarrow R \rightarrow y \rightarrow \hat{y},
\end{equation}
which explicitly models evidence extraction, stepwise reasoning, and consistency-aware decision making. This formulation provides a unified basis for integrating domain guidelines with computational reasoning, and serves as the foundation for the proposed framework.

\section{Methodology: G-SHARE Framework}\label{sec:method}

\subsection{Framework overview}

This section presents the overall architecture of the proposed G-SHARE framework, which is designed to operationalize the diagnostic procedure into a structured and reproducible reasoning pipeline. Rather than performing direct text-to-label mapping, the framework decomposes the human-factor event diagnosis task into a sequence of interconnected stages, each corresponding to a distinct analytical function.

As illustrated in Fig.~\ref{fig:framework_overview2}, the framework takes a human-factor event report $x = (b, s)$ as input and produces a final diagnosis $\hat{y}$ through three main stages: evidence extraction, stepwise guideline-based diagnosis, and logical consistency repair. This staged design explicitly models the intermediate reasoning process and ensures that the final output is both interpretable and aligned with the underlying diagnostic logic.

\begin{center}
\includegraphics[width=1.0 \textwidth]{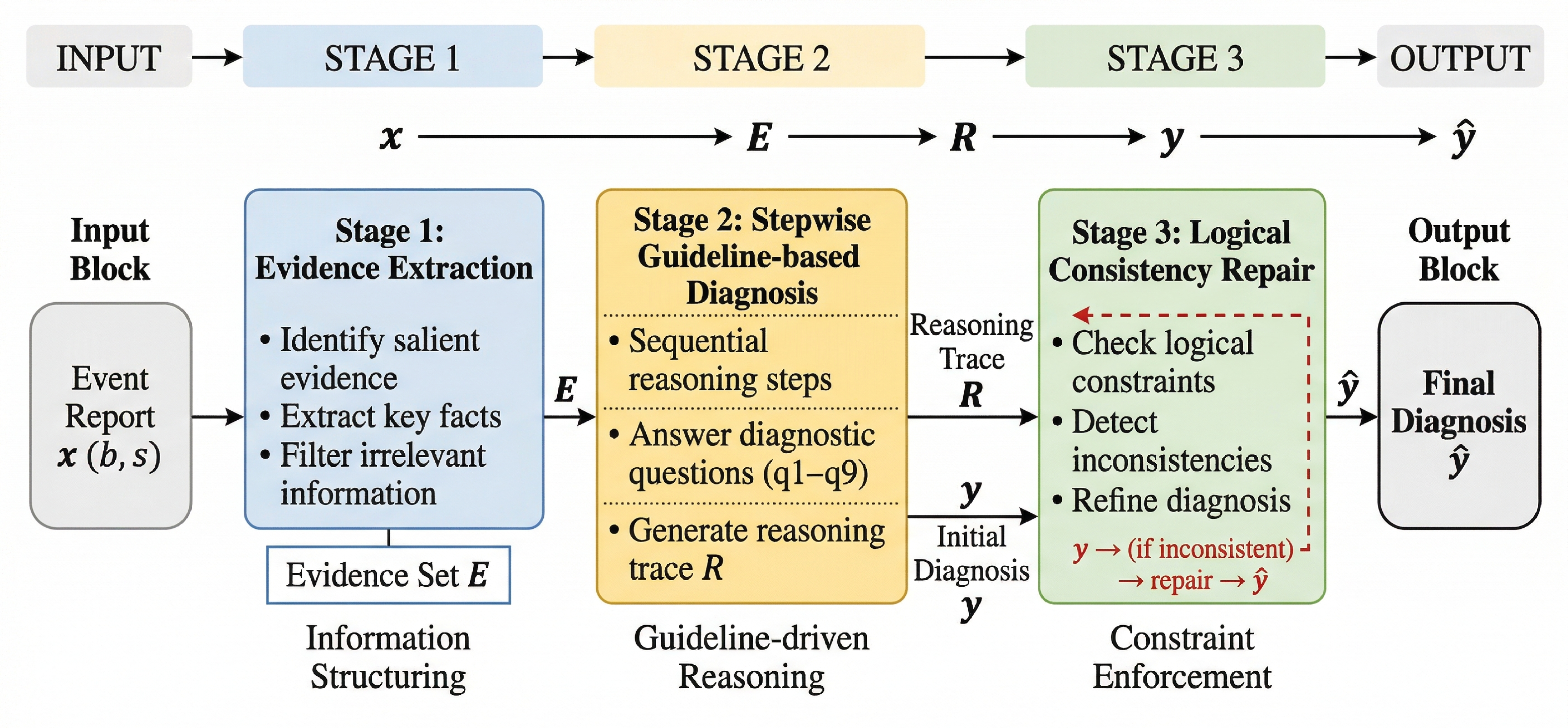}
\captionof{figure}{Overall architecture of the G-SHARE framework.}
\label{fig:framework_overview2}
\end{center}

In the first stage, \emph{evidence extraction}, the framework identifies and structures salient information from the input report, producing an evidence set $E$. This step aims to filter irrelevant content and highlight key elements such as operator actions, system states, and contextual conditions, thereby providing a structured basis for subsequent reasoning. In the second stage, \emph{stepwise guideline-based diagnosis}, the framework generates a reasoning trace $R$ by sequentially answering a set of diagnostic questions derived from the guideline. Each step corresponds to an intermediate judgment, and together they form a structured interpretation of the event. Based on this reasoning trace, an initial diagnosis $y$ is obtained through a predefined decision structure. In the third stage, \emph{logical consistency repair}, the framework evaluates the consistency between intermediate reasoning steps and the initial diagnosis. When inconsistencies or logical conflicts are detected, a refinement process is applied to produce a corrected diagnosis $\hat{y}$. This stage acts as a safeguard mechanism to ensure that the final output adheres to the constraints implied by the diagnostic procedure.

Overall, the proposed framework can be summarized as a structured mapping:
\begin{equation}
    x \rightarrow E \rightarrow R \rightarrow y \rightarrow \hat{y},
\end{equation}
which explicitly captures evidence extraction, reasoning generation, and consistency-aware decision making. By decomposing the diagnosis process into well-defined stages, G-SHARE provides a transparent and reproducible pipeline that facilitates both interpretability and reliability in human-factor event analysis.

\subsection{Stage 1: Evidence extraction}

The first stage of the proposed framework aims to extract salient evidence from the input event report and construct a structured representation to support subsequent reasoning. Rather than summarizing the report, this stage focuses on identifying diagnostically relevant information and filtering out irrelevant or redundant content.

Given an event report $x = (b, s)$, the objective of evidence extraction is to produce an evidence set $E = \{e_1, e_2, \dots, e_m\}$, where each element $e_i$ corresponds to a key fact or observation derived from the report. The extraction targets include, but are not limited to: (i) operator actions and behavioral descriptions, (ii) system states and operational conditions, (iii) temporal relationships between events, and (iv) contextual factors that may influence human performance. These elements are selected based on their relevance to the diagnostic procedure and their potential contribution to subsequent reasoning steps.

The output of this stage is a structured evidence set $E$, which can be represented as a collection of concise textual statements or structured tuples. Compared with the original narrative report, $E$ provides a more focused and organized view of the information that is directly relevant to diagnosis. This representation is intended to serve as an intermediate layer between unstructured input and structured reasoning, enabling more controlled and interpretable downstream processing.

From a methodological perspective, this stage is motivated by the observation that human analysts do not rely on the full narrative text during diagnosis, but instead selectively attend to key pieces of evidence. By explicitly modeling this process, the framework aims to improve evidence salience and reduce the impact of irrelevant information, thereby supporting more stable and consistent reasoning in subsequent stages.

It is important to note that the effectiveness of evidence extraction may depend on multiple factors, including report length, information density, prompt design, and the capability of the underlying language model. In scenarios where reports are relatively concise or where the model is capable of implicitly identifying relevant information, the marginal benefit of this stage may be limited. Nevertheless, the explicit construction of an evidence set provides a structured interface for reasoning and facilitates interpretability and potential extensions, such as human-in-the-loop verification or evidence-level analysis.

\subsection{Stage 2: Guideline-based structured reasoning}

The second stage constitutes the core of the proposed framework, where the diagnostic procedure is implemented as a structured reasoning process. Unlike free-form generation, this stage explicitly follows a predefined sequence of diagnostic steps derived from the guideline, ensuring that the reasoning process is aligned with expert analytical practice.

Given the evidence set $E$, the objective of this stage is to construct a reasoning trace $R = (r_1, r_2, \dots, r_k)$, where each element $r_i$ corresponds to an intermediate judgment associated with a diagnostic question. The reasoning trace serves as an explicit representation of the decision-making process, enabling both interpretability and verification of intermediate steps.

\subsubsection{Mapping the guideline to reasoning steps}

The original diagnostic guideline consists of a sequence of interdependent analytical steps, each addressing a specific aspect of event interpretation. To enable computational implementation, these steps are translated into a set of structured reasoning operations, where each step is associated with a well-defined input, output, and decision scope.

Specifically, each diagnostic question $q_i$ is mapped to a reasoning step that produces an answer $a_i \in \mathcal{A}_i$, based on the available evidence $E$ and previously inferred answers $\{a_1, \dots, a_{i-1}\}$. This formulation captures the sequential dependency inherent in the diagnostic procedure and ensures that later judgments are conditioned on earlier interpretations.

During this mapping process, certain guideline steps are combined or simplified to improve computational tractability. For example, closely related sub-steps that involve similar evidence interpretation may be merged into a single reasoning operation, while implicit expert knowledge embedded in the guideline is translated into explicit decision criteria. Despite these transformations, the overall logical structure of the original procedure is preserved, maintaining consistency with expert practice.

As a result, the reasoning process can be expressed as a sequential inference:
\begin{equation}
    a_i = f_i(E, a_1, \dots, a_{i-1}), \quad i = 1, \dots, k,
\end{equation}
where each function $f_i(\cdot)$ represents a structured reasoning step. The final diagnosis is then determined by applying a decision function to the complete reasoning trace:
\begin{equation}
    y = f_{\text{dec}}(a_1, a_2, \dots, a_k).
\end{equation}

\subsubsection{Prompting strategy and output schema}

To operationalize the structured reasoning process, each step is implemented using controlled prompting that specifies the diagnostic question, relevant evidence, and expected output format. Unlike free-form prompting, the model is guided to produce constrained responses that correspond to predefined answer spaces, reducing ambiguity and improving consistency.

At each step, the model receives as input the evidence set $E$, the current diagnostic question $q_i$, and optionally the previously generated answers. The output is required to include both a selected answer $a_i$ and a brief rationale that explains the decision based on the available evidence. This design ensures that the reasoning trace remains interpretable and can be inspected or validated if necessary.

The complete reasoning trace is represented in a structured format, such as a sequence of key-value pairs or a JSON-like schema:
\begin{equation}
    R = \{(q_1, a_1, r_1), (q_2, a_2, r_2), \dots, (q_k, a_k, r_k)\},
\end{equation}
where each tuple contains the diagnostic question, the selected answer, and the corresponding rationale. This representation enables explicit tracking of intermediate decisions and facilitates downstream consistency checking.

Based on this design, two variants of the reasoning strategy are considered. The first variant (V1) implements a basic stepwise diagnosis, where each question is answered sequentially using a consistent prompting template. The second variant (V2) enhances this process by introducing more explicit instructions, stronger constraints on answer selection, and clearer alignment between evidence and rationale. In particular, V2 emphasizes precise category discrimination and requires that each answer be explicitly supported by identified evidence, thereby improving the robustness and interpretability of the reasoning trace.

Overall, this stage transforms the diagnostic procedure into a machine-executable reasoning workflow that preserves the stepwise logic of expert analysis while enabling transparent and auditable decision making.

\subsection{Stage 3: Logic-based consistency repair}

While the structured reasoning stage enforces stepwise decision making, the resulting outputs may still exhibit inconsistencies due to uncertainty in intermediate judgments or limitations of the underlying language model. In safety-critical human-factor diagnosis, it is not sufficient for the final result to be plausible; it must also be logically consistent with the diagnostic procedure and the intermediate reasoning steps. To address this requirement, a logic-based consistency repair mechanism is introduced as the third stage of the framework.

In this context, \emph{consistency} refers to the alignment between intermediate answers in the reasoning trace $R = \{(q_i, a_i, r_i)\}$ and the final diagnosis $y$, as well as the internal coherence among the answers themselves. Specifically, the reasoning trace must satisfy the logical constraints implied by the diagnostic procedure, including conditional dependencies between steps, mutual exclusivity of certain answer combinations, and compatibility between inferred mechanisms and assigned categories.

To formalize this requirement, a set of consistency rules $\mathcal{C}$ is defined based on the decision structure of the diagnostic procedure. These rules capture key logical relationships, such as: (i) prerequisite constraints, where the validity of a step depends on the outcome of previous steps; (ii) implication constraints, where certain combinations of answers necessarily lead to specific diagnostic outcomes; and (iii) exclusion constraints, where incompatible answers or categories are not allowed to co-occur. Given a reasoning trace $R$ and an initial diagnosis $y$, consistency checking can be expressed as a validation function:
\begin{equation}
    \text{Consistent}(R, y) = 
    \begin{cases}
        1, & \text{if all constraints in } \mathcal{C} \text{ are satisfied}, \\
        0, & \text{otherwise}.
    \end{cases}
\end{equation}

When inconsistencies are detected, a repair process is triggered to produce a corrected diagnosis $\hat{y}$. The repair mechanism operates by re-evaluating the relationship between the reasoning trace and the decision rules. In its simplest form, this can be achieved by reapplying the deterministic decision function $f_{\text{dec}}(\cdot)$ to the validated subset of answers in $R$, or by adjusting conflicting answers to satisfy the constraints before recomputing the final outcome. In practice, the repair process prioritizes preserving evidence-supported intermediate judgments while enforcing consistency with the decision structure.

The refined diagnosis is then given by
\begin{equation}
    \hat{y} = f_{\text{repair}}(R, y, \mathcal{C}),
\end{equation}
where $f_{\text{repair}}(\cdot)$ denotes the consistency-aware correction function. This function ensures that the final output adheres to the logical requirements of the diagnostic procedure, even when the initial reasoning trace contains minor inconsistencies.

This module acts as a lightweight safety layer for diagnosis outputs. By introducing explicit constraint checking and repair, the framework enhances robustness against reasoning errors and improves the trustworthiness of the diagnostic results, which is particularly important in safety-critical applications.

\subsection{Design rationale}

The design of G-SHARE is motivated by the requirements of human-factor event diagnosis in safety-critical environments, where transparency, consistency, and alignment with expert procedures are as important as predictive performance. This section explains the key design choices of the framework from a methodological perspective.

\paragraph{Limitations of direct text-to-diagnosis mapping}
Direct one-shot or free-form large language model approaches map an input report $x$ to a diagnosis $y$ through unconstrained generation. Although such methods can produce plausible outputs, they provide limited control over the reasoning process and offer no explicit guarantees regarding consistency with domain procedures. In human-factor diagnosis, conclusions must be grounded in evidence and aligned with predefined decision logic. The lack of structured reasoning therefore limits their suitability for safety-critical applications.

\paragraph{Role of structured, stepwise reasoning}
The diagnostic procedure inherently follows a sequential logic, where intermediate judgments condition subsequent decisions. To reflect this property, the framework constructs an explicit reasoning trace $R$, in which each step corresponds to a diagnostic question and produces an evidence-supported answer. This design transforms the task into a structured inference process:
\[
x \rightarrow E \rightarrow R \rightarrow y,
\]
enabling interpretability at the step level and improving alignment with expert analytical workflows.

\paragraph{Need for logic-based consistency enforcement}
Even with structured reasoning, inconsistencies may arise due to uncertainty in intermediate judgments or limitations of the underlying model. In safety-critical contexts, such inconsistencies must be explicitly addressed. The introduction of a consistency repair stage enforces logical constraints derived from the diagnostic procedure and produces a refined output $\hat{y}$:
\[
y \rightarrow \hat{y}.
\]
This mechanism ensures coherence between intermediate reasoning steps and final decisions, enhancing robustness and reliability.

\paragraph{Alignment with safety-critical application requirements}
The primary objective of the proposed framework is not to maximize benchmark performance, but to provide a trustworthy and auditable diagnostic workflow. In safety-critical domains, decision support systems must support traceability, enable human verification, and adhere to established analytical procedures. By integrating evidence structuring, guideline-aligned reasoning, and constraint-based validation, G-SHARE provides a structured pipeline that satisfies these requirements and supports practical deployment.

Overall, the design reflects a transition from unconstrained prediction to structured, procedure-aware reasoning, emphasizing reliability and interpretability as core objectives.

\section{Experimental Design}\label{sec:Experimental Design}

The experimental design evaluates the proposed framework from three perspectives: the necessity of structured reasoning over direct classification, the contribution of constraint enforcement to logical consistency and accuracy, and the advantage of LLM-based semantic understanding over traditional machine learning.

\subsection{Dataset construction}\label{sec:dataset}

The dataset used in this study consists of 266 real-world human-factor event reports collected from operational nuclear power plant records in China. Each report includes a behavioral descriptor $b$ and a narrative summary $s$, with lengths ranging from approximately 30 to 120 Chinese characters. These reports document operator actions, system responses, and contextual conditions associated with abnormal events. The diagnosis follows a structured nine-step procedure, which serves as the basis for the proposed reasoning framework.

To ensure relevance to the diagnostic task, reports were selected based on the presence of sufficient human-factor information, including explicit descriptions of operator behavior, system state transitions, and contextual factors. Reports lacking human involvement or containing insufficient narrative detail were excluded.

\paragraph{Class distribution.}
Table~\ref{tab:class_dist} and Fig.~\ref{fig:class_distribution_sci_top} present the class distribution of the gold-standard subset. The distribution is highly imbalanced, with the majority of cases belonging to the \textit{Violation} category, while several categories are not represented.

\begin{center}
\includegraphics[width=0.6\textwidth]{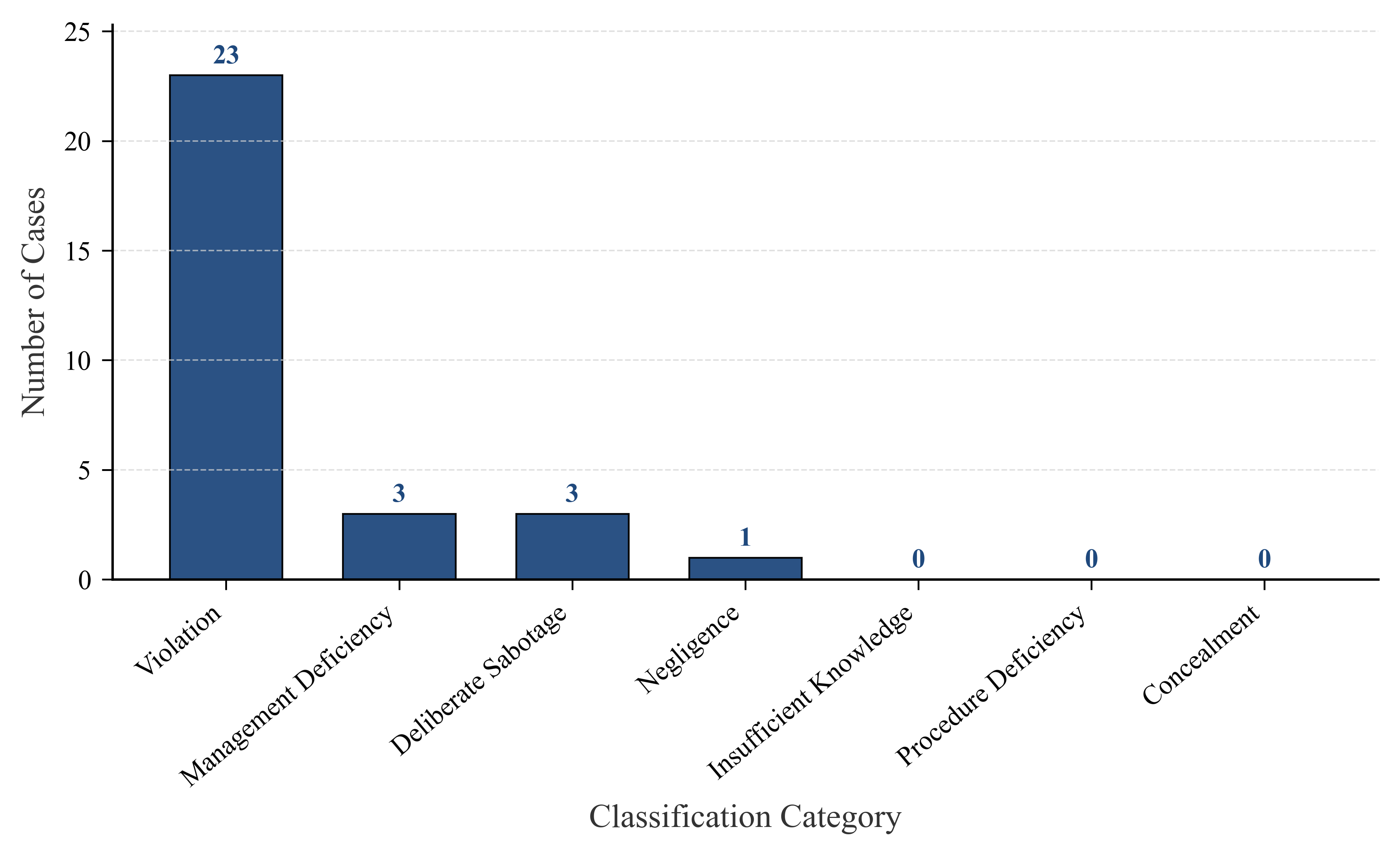}
\captionof{figure}{Class distribution of the gold-standard dataset.}
\label{fig:class_distribution_sci_top}
\end{center}

As shown in Fig.~\ref{fig:class_distribution_sci_top}, the dataset is dominated by violation-related events, accounting for the majority of annotated cases, whereas several categories have zero instances. This imbalance reflects the inherent rarity and uneven distribution of human-factor event types in real operational environments, rather than an artifact of dataset construction. In practice, deviations from established procedures constitute the most common type of human-factor event, while categories such as knowledge deficiency or concealment occur infrequently.

\begin{table}[htbp]
\centering
\caption{Gold-standard class distribution (30 expert-annotated cases). Several categories are not represented due to their rarity in operational data.}
\label{tab:class_dist}
\begin{tabular}{lcc}
\toprule
\textbf{Class} & \textbf{Count} & \textbf{Percentage} \\
\midrule
Violation & 23 & 76.7\% \\
Management Deficiency & 3 & 10.0\% \\
Deliberate Sabotage & 3 & 10.0\% \\
Negligence & 1 & 3.3\% \\
Insufficient Knowledge & 0 & 0.0\% \\
Procedure Deficiency & 0 & 0.0\% \\
Concealment & 0 & 0.0\% \\
\midrule
\textbf{Total} & \textbf{30} & \textbf{100\%} \\
\bottomrule
\end{tabular}
\end{table}

Although such imbalance poses challenges for conventional classification models, the proposed framework mitigates this issue by leveraging structured reasoning and guideline-based decision processes, which reduce reliance on purely statistical patterns and enable more robust interpretation under sparse data conditions.

\paragraph{Gold-standard annotation.}
A gold-standard subset of 30 reports was constructed for evaluation. The annotation was performed by a domain expert with more than 15 years of experience in nuclear safety analysis. Each report was annotated following the structured diagnostic procedure, including both the final category label and the answers to all intermediate decision steps, resulting in complete decision trajectories.

To improve annotation quality, a stratified disagreement sampling strategy was adopted, prioritizing cases where preliminary model outputs or heuristic rules exhibited uncertainty or disagreement. This approach focuses expert effort on diagnostically challenging cases and enhances the informativeness of the evaluation set. All annotations were reviewed for internal consistency with the diagnostic procedure.

\paragraph{Decision trajectory representation.}
In addition to final labels, each annotated case includes a complete sequence of intermediate decisions. Table~\ref{tab:trajectory_example} presents an example of such a decision trajectory, illustrating how a diagnosis is derived through stepwise reasoning.

\begin{table*}[htbp]
\centering
\caption{Example of a decision trajectory following the structured diagnostic procedure.}
\label{tab:trajectory_example}
\begin{tabular}{cccc}
\toprule
\textbf{Step} & \textbf{Question} & \textbf{Answer} & \textbf{Rationale (abridged)} \\
\midrule
$q_1$ & Intentional behavior? & No & No evidence of deliberate intent \\
$q_2$ & Awareness of consequence? & Yes & Operator acknowledged abnormal condition \\
$q_3$ & Procedure exists? & Yes & Relevant procedure documented \\
$q_4$ & Procedure correct? & Yes & No defect identified in procedure \\
$q_5$ & Universal issue? & No & Specific to this case \\
$q_6$ & Knowledge adequate? & Yes & Operator trained for the task \\
$q_7$ & Historical occurrence? & No & No prior similar events \\
$q_8$ & Other misconduct? & No & No violation beyond procedure deviation \\
$q_9$ & Reported truthfully? & Yes & Event reported according to protocol \\
\midrule
\multicolumn{4}{c}{\textbf{Final Diagnosis: Violation}} \\
\bottomrule
\end{tabular}
\end{table*}

The availability of step-level annotations enables a more detailed evaluation of reasoning processes, beyond final classification accuracy. This structured representation aligns with the design of the proposed framework and supports analysis of intermediate decision quality.

\subsection{Baselines and compared methods}\label{sec:baselines}

To evaluate the effectiveness of the proposed framework, we compare G-SHARE with representative approaches from three categories: (i) free-form LLM-based diagnosis, (ii) structured reasoning variants, and (iii) traditional machine learning methods. All methods take the same input $(b, s)$ for fair comparison. The purpose of this comparison is not merely to identify the strongest classifier, but to examine whether guideline alignment and structured reasoning improve safety-oriented diagnosis.

\paragraph{Free-form LLM baselines.}
The first category includes direct LLM-based diagnosis without explicit reasoning structure. In this setting, the model predicts one of the seven diagnostic categories in a single step, based solely on the input report. This one-shot baseline reflects unconstrained generation and serves as a reference for evaluating the impact of structured reasoning.

\paragraph{Structured reasoning variants.}
The second category includes stepwise reasoning approaches derived from the diagnostic procedure. In the stepwise baseline without constraint enforcement, the reasoning process follows the sequential diagnostic steps, but the final decision is obtained directly from intermediate answers without applying logical consistency repair. The proposed G-SHARE framework is evaluated under two variants. G-SHARE V1 employs a basic prompting strategy with a fixed stepwise template and post-hoc constraint repair. G-SHARE V2 introduces enhanced instruction specificity, stronger category discrimination, and explicit rationale anchoring, while maintaining the same repair mechanism. The comparison between these variants isolates the effect of structured reasoning design and constraint enforcement.

\paragraph{Traditional machine learning baselines.}
The third category includes conventional text classification methods. We implement a Logistic Regression model with TF-IDF features constructed from the concatenated report text, using jieba tokenization and one-vs-rest classification with balanced class weights. Evaluation is conducted using leave-one-out cross-validation on the 30 gold-standard cases.

\paragraph{LLM models evaluated.}
For the structured reasoning framework, multiple API-based large language models are evaluated to assess robustness across model backbones, including DeepSeek-V3.2 \cite{deepseek2024}, Qwen2.5-72B-Instruct (Qwen-72B) \cite{qwen2025}, Qwen3-80B-Instruct (Qwen-80B), and Qwen3-80B-Thinking (Qwen-80B-T). All models are tested under consistent settings to ensure comparability.

\subsection{Evaluation metrics}\label{sec:metrics}

Performance is evaluated using both outcome-level and process-level metrics.

\paragraph{Primary metrics.}
Accuracy and Macro-F1\cite{hinojosa2024performance} are reported as the main evaluation metrics. Accuracy measures the proportion of correctly classified cases. Macro-F1 is computed as the unweighted average of F1 scores across all classes. Given the class imbalance in the dataset, Macro-F1 provides a more balanced assessment by accounting for minority categories.

\paragraph{Per-class performance.}
Per-class recall is used to examine performance across different diagnostic categories, particularly those with limited representation. Detailed results are provided in the supplementary material.

\paragraph{Step-level accuracy.}
Step accuracy evaluates the correctness of intermediate decisions in the reasoning trace $R$. It is defined as the proportion of correctly predicted answers $a_i$ across all diagnostic steps. This metric enables identification of error propagation within the sequential reasoning process.

\paragraph{Logical consistency.}
The logical inconsistency rate is defined as the proportion of cases in which the reasoning trace and final diagnosis violate predefined decision constraints. This metric is used to quantify the effect of constraint enforcement in ablation studies.

\paragraph{Error-origin analysis.}
For misclassified cases, the first divergent step is identified as the earliest step at which the predicted answer differs from the reference trajectory. This analysis characterizes whether errors originate from early-stage evidence interpretation or later-stage decision inconsistency.

\subsection{Results}\label{sec:results}

\subsubsection{Overall diagnostic performance}

Table~\ref{tab:main_comparison} presents the overall diagnostic performance of all compared methods on the 30 gold-standard cases, evaluated using both accuracy and Macro-F1.

\begin{table}[htbp]
\centering
\caption{Overall comparison on 30 gold-standard cases. Best results are shown in bold.}
\label{tab:main_comparison}
\begin{tabular}{lcc}
\toprule
\textbf{Method} & \textbf{Accuracy} & \textbf{Macro-F1} \\
\midrule
Logistic Regression + TF-IDF & 76.7\% & 0.341 \\
One-shot LLM (Qwen-80B) & 56.7\% & 0.341 \\
Stepwise V1, no constraint (Qwen-80B) & 3.3\% & 0.125 \\
Stepwise V2, no constraint (Qwen-80B) & 83.3\% & 0.605 \\
G-SHARE V1 (Qwen-80B) & 50.0\% & 0.304 \\
\textbf{G-SHARE V2 (Qwen-80B)} & \textbf{83.3\%} & \textbf{0.605} \\
\bottomrule
\end{tabular}
\end{table}

Among all methods, G-SHARE V2 achieves the best overall performance, reaching an accuracy of 83.3\% and a Macro-F1 of 0.605. This result indicates that the proposed framework can effectively capture both dominant and minority classes under a highly imbalanced data distribution.

Compared with one-shot large language model inference, which attains 56.7\% accuracy and a Macro-F1 of 0.341, the structured reasoning approach leads to substantial improvements. The relatively low performance of one-shot diagnosis suggests that direct text-to-label mapping is insufficient for this task, as it lacks explicit control over intermediate reasoning and often fails to utilize diagnostically relevant evidence in a consistent manner. Traditional machine learning methods based on TF-IDF features achieve a relatively high accuracy of 76.7\%, but their Macro-F1 remains low (0.341), indicating poor performance on minority classes. This reflects a tendency to overfit the dominant category in imbalanced datasets. In contrast, G-SHARE maintains a significantly higher Macro-F1 while achieving comparable or superior accuracy, demonstrating more balanced performance across categories. The comparison between stepwise variants further highlights the importance of structured design. While the unconstrained stepwise V1 model performs poorly, the improved V2 variant achieves strong results, indicating that careful formulation of reasoning steps and decision constraints is critical for effective diagnosis.

Overall, the superiority of G-SHARE suggests that structured, guideline-aligned reasoning is more effective than unconstrained generation for human-factor diagnosis, particularly in safety-critical scenarios with limited and imbalanced data.

\subsubsection{Per-Class Performance Analysis}

Table~\ref{tab:per_class_recall} reports per-class recall on the 30 gold-standard cases. G-SHARE V2 achieves perfect recall on the dominant Violation class and correctly identifies the single Negligence case, demonstrating strong performance on well-defined behavioral categories. However, performance on minority classes remains limited. In particular, Management Deficiency is not correctly identified by any method, while Deliberate Sabotage achieves only partial recall.

\begin{table}[htbp]
\centering
\caption{Per-class recall on 30 gold-standard cases. ``---'' indicates that no gold cases exist for that class.}
\label{tab:per_class_recall}
\resizebox{\linewidth}{!}{%
\begin{tabular}{lcccccc}
\toprule
\textbf{Class} & \textbf{Gold} & \textbf{LR+TF-IDF} & \textbf{One-shot} & \textbf{No-Constr.\ V2} & \textbf{G-SHARE V1} & \textbf{G-SHARE V2} \\
\midrule
Violation (23) & 23 & \textbf{0.957} & 0.652 & \textbf{1.000} & 0.609 & \textbf{1.000} \\
Mgmt.\ Deficiency (3) & 3 & 0.000 & 0.333 & 0.000 & 0.000 & \textbf{0.000} \\
Delib.\ Sabotage (3) & 3 & 0.333 & 0.000 & 0.333 & 0.333 & \textbf{0.333} \\
Negligence (1) & 1 & 0.000 & \textbf{1.000} & \textbf{1.000} & 0.000 & \textbf{1.000} \\
Insuff.\ Knowledge & --- & --- & --- & --- & --- & --- \\
Proc.\ Deficiency & --- & --- & --- & --- & --- & --- \\
Concealment & --- & --- & --- & --- & --- & --- \\
\bottomrule
\end{tabular}%
}
\end{table}

It is important to note that the dataset exhibits severe class imbalance, with 23 out of 30 cases belonging to the Violation category, while minority classes contain only one to three instances. As a result, per-class recall for these categories is highly sensitive to individual prediction errors and should be interpreted with caution.

The consistently low recall for Management Deficiency across all methods suggests that this category is intrinsically difficult, as it often requires recognizing implicit procedure defects rather than explicit operator actions. This observation is further supported by the error analysis in Section~\ref{sec:error_analysis}, where misclassifications concentrate on procedure-related ambiguity.

\subsubsection{Statistical Significance}

\paragraph{Bootstrap confidence intervals.}
Table~\ref{tab:bootstrap_ci} presents 95\% bootstrap confidence intervals based on 10{,}000 resamples.

G-SHARE V2 achieves the highest performance with an accuracy of 83.5\% [70.0, 96.7] and a Macro-F1 of 0.547 [0.295, 0.750]. Compared with baseline methods, the confidence intervals indicate a clear improvement over one-shot LLM and G-SHARE V1, while partially overlapping with LR+TF-IDF due to the latter's strong bias toward the dominant class.

\paragraph{McNemar's test.}
Table~\ref{tab:mcnemar} reports paired significance tests between G-SHARE V2 and baseline methods.

\begin{table}[htbp]
\centering
\caption{McNemar's test: G-SHARE V2 vs.\ baselines (30 gold cases).}
\label{tab:mcnemar}
\begin{tabular}{lcccc}
\toprule
\textbf{Comparison} & $b$ & $c$ & $\chi^2$ & $p$-value \\
\midrule
G-SHARE V2 vs.\ LR+TF-IDF & 3 & 1 & 0.25 & 0.617 \\
G-SHARE V2 vs.\ One-shot LLM & 9 & 1 & 4.90 & 0.027\textsuperscript{*} \\
G-SHARE V2 vs.\ G-SHARE V1 & 10 & 0 & 8.10 & 0.004\textsuperscript{**} \\
\bottomrule
\end{tabular}
\end{table}

The comparison with one-shot LLM is statistically significant ($p = 0.027$), and the improvement over G-SHARE V1 is highly significant ($p = 0.004$), confirming the effectiveness of structured reasoning design. In contrast, the difference with LR+TF-IDF is not statistically significant ($p = 0.617$), which is consistent with its high accuracy on the dominant class despite poor minority-class performance.

\subsubsection{Effect of Structured Reasoning Design}

Table~\ref{tab:label_acc} summarizes the performance improvement from V1 to V2 across different models. For all fully evaluated models, V2 consistently outperforms V1, with Qwen-80B achieving the largest gain (+33.3\% accuracy and +0.301 Macro-F1). Similar improvements are observed for Qwen-72B, indicating that the performance gain is robust across model variants.

\begin{table}[htbp]
\centering
\caption{Label accuracy and Macro-F1 on the 30 gold-standard cases across models. $\Delta$ denotes improvement from V1 to V2.}
\label{tab:label_acc}
\resizebox{\linewidth}{!}{%
\begin{tabular}{lcccccc}
\toprule
& \multicolumn{3}{c}{\textbf{Accuracy}} & \multicolumn{3}{c}{\textbf{Macro-F1}} \\
\cmidrule(lr){2-4} \cmidrule(lr){5-7}
\textbf{Model} & V1 & V2 & $\Delta$ & V1 & V2 & $\Delta$ \\
\midrule
DeepSeek-V3.2 & 100.0\%\textsuperscript{a} & --- & --- & 1.000\textsuperscript{a} & --- & --- \\
Qwen-72B & 48.3\% & 76.7\% & +28.4 & 0.339 & 0.569 & +0.230 \\
Qwen-80B & 50.0\% & \textbf{83.3\%} & +33.3 & 0.304 & \textbf{0.605} & +0.301 \\
Qwen-80B-T & 73.9\%\textsuperscript{b} & --- & --- & 0.399\textsuperscript{b} & --- & --- \\
\bottomrule
\multicolumn{7}{l}{\footnotesize \textsuperscript{a} DeepSeek V1 covered only 9 of 30 gold cases and is not directly comparable.} \\
\multicolumn{7}{l}{\footnotesize \textsuperscript{b} Qwen-80B-T V1 covered 23 of 30 gold cases.}
\end{tabular}%
}
\end{table}

As shown in Fig.~\ref{fig:v1_v2_improvement}, the transition from V1 to V2 consistently improves performance across models, with substantial gains observed in both accuracy and Macro-F1. Notably, the magnitude of improvement exceeds +30\% for Qwen-80B, indicating a significant enhancement rather than marginal tuning. Since both V1 and V2 use the same underlying models, this result confirms that the observed performance gain is driven by improvements in reasoning design rather than model capacity.

\begin{center}
\includegraphics[width=0.7 \textwidth]{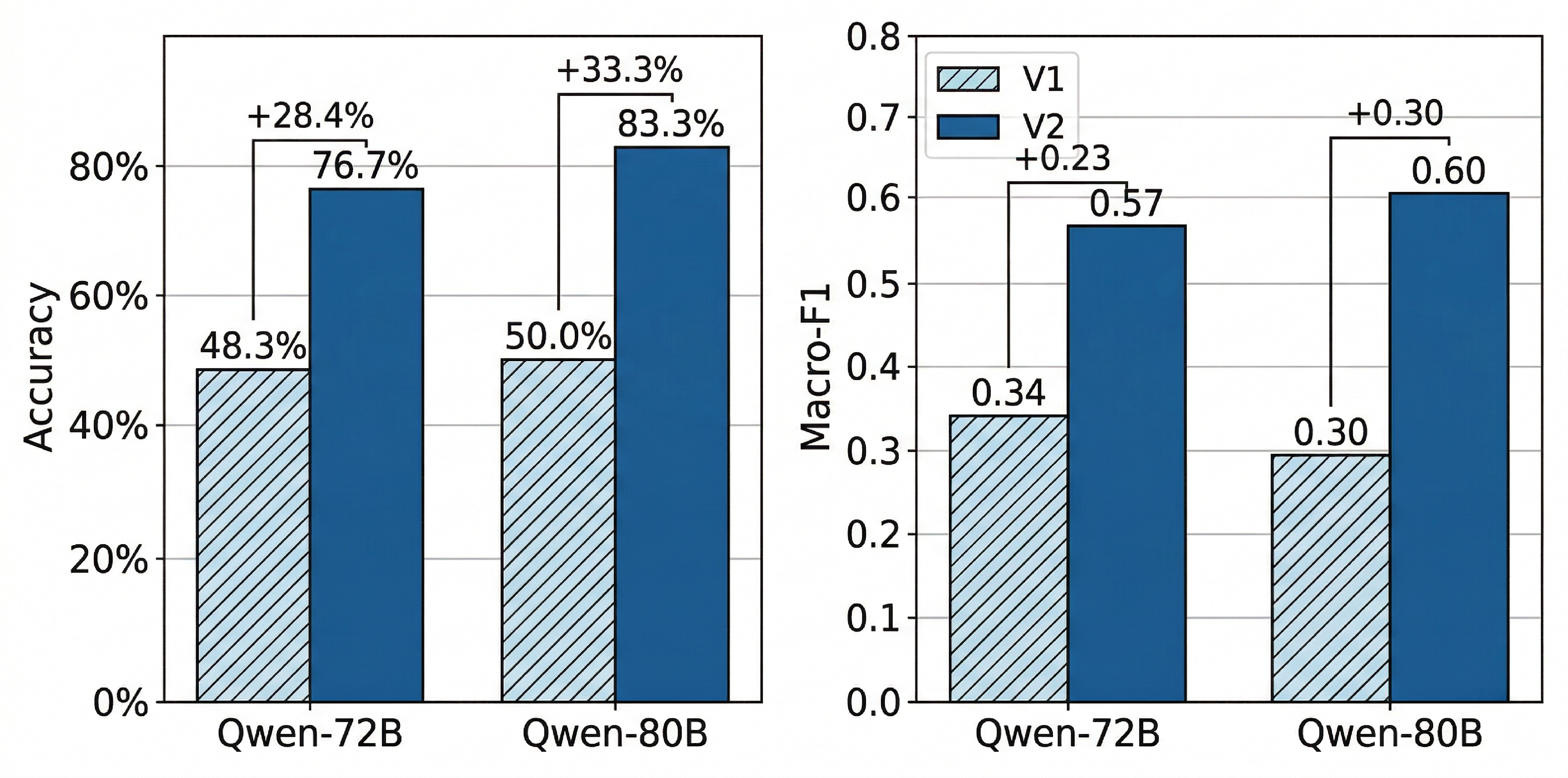}
\captionof{figure}{Performance improvement from V1 to V2 across models.}
\label{fig:v1_v2_improvement}
\end{center}

To further understand the source of improvement, Table~\ref{tab:step_acc} reports step-level accuracy. The largest gains are observed in Step~3 (procedure existence), Step~4 (procedure correctness), and Step~8 (other misconduct), with improvements exceeding +30\% in some cases. These steps correspond to semantically ambiguous decision points where the distinction between categories (e.g., absence vs.\ violation of procedures) is subtle and often implicit in the narrative. The results indicate that the refined prompt effectively resolves these ambiguities by enforcing clearer decision boundaries.

\begin{table}[htbp]
\centering
\caption{Step-level accuracy (\%) on gold-standard cases: V1 vs.\ V2. Bold indicates the higher value.}
\label{tab:step_acc}
\resizebox{\linewidth}{!}{%
\begin{tabular}{clcccccc}
\toprule
& & \multicolumn{3}{c}{\textbf{Qwen-80B}} & \multicolumn{3}{c}{\textbf{Qwen-72B}} \\
\cmidrule(lr){3-5} \cmidrule(lr){6-8}
\textbf{Step} & \textbf{Decision} & V1 & V2 & $\Delta$ & V1 & V2 & $\Delta$ \\
\midrule
1 & Intentional? & 73.3 & \textbf{76.7} & +3.3 & \textbf{82.8} & 80.0 & $-$2.8 \\
2 & Aware of consequence? & 80.0 & \textbf{83.3} & +3.3 & 58.6 & \textbf{66.7} & +8.1 \\
3 & Procedure exists? & 60.0 & \textbf{96.7} & +36.7 & 62.1 & \textbf{93.3} & +31.3 \\
4 & Procedure correct? & 60.7 & \textbf{92.9} & +32.1 & 64.3 & \textbf{89.3} & +25.0 \\
5 & Error universal? & 66.7 & \textbf{76.7} & +10.0 & 72.4 & \textbf{76.7} & +4.3 \\
6 & Training deficient? & 58.3 & \textbf{75.0} & +16.7 & \textbf{69.6} & 66.7 & $-$2.9 \\
7 & Historical behavior? & 78.9 & 78.9 & 0.0 & \textbf{77.8} & 73.7 & $-$4.1 \\
8 & Other misconduct? & 33.3 & \textbf{96.7} & +63.3 & 65.5 & \textbf{100.0} & +34.5 \\
9 & Reported? & 100.0 & 50.0 & $-$50.0 & 100.0 & 100.0 & 0.0 \\
\midrule
\multicolumn{2}{l}{\textbf{Macro Average}} & 67.9 & \textbf{80.8} & +12.8 & 72.6 & \textbf{82.9} & +10.4 \\
\bottomrule
\end{tabular}%
}
\end{table}

Further evidence is provided by the prompt component ablation in Table~\ref{tab:prompt_ablation}. Among different components, disambiguation rules contribute the most significant performance gain, whereas distributional priors and anchoring strategies alone yield only moderate improvements. This finding suggests that explicitly encoding domain-specific distinctions within the reasoning process is more critical than providing general statistical guidance.

\begin{table}[htbp]
\centering
\caption{Prompt component ablation (Qwen-80B, with evidence extraction and constraint repair).}
\label{tab:prompt_ablation}
\begin{tabular}{lcc}
\toprule
\textbf{Prompt Variant} & \textbf{Accuracy} & \textbf{Macro-F1} \\
\midrule
V1 (baseline) & 50.0\% & 0.300 \\
+ Distributional priors & 63.3\% & 0.385 \\
+ Disambiguation rules & 83.3\% & 0.475 \\
+ Priors + disambiguation & 80.0\% & 0.350 \\
+ All components (priors + disambig.\ + anchoring) & \textbf{80.0\%} & \textbf{0.350} \\
\midrule
G-SHARE V2 (full, original run) & \textbf{83.3\%} & \textbf{0.605} \\
\bottomrule
\end{tabular}
\end{table}

In addition, the distribution shift observed in Table~\ref{tab:dist_shift} shows that V2 substantially reduces the over-prediction of rare categories and aligns the predicted label distribution with domain expectations. This indicates that structured reasoning not only improves accuracy on individual cases but also enhances global consistency at the dataset level.

\begin{table}[htbp]
\centering
\caption{Predicted label distribution over 266 cases (Qwen-80B). Both settings include constraint repair.}
\label{tab:dist_shift}
\begin{tabular}{lrr}
\toprule
\textbf{Label} & \textbf{V1} & \textbf{V2} \\
\midrule
Violation & 144 & 262 \\
Insufficient Knowledge & 64 & 2 \\
Negligence & 41 & 1 \\
Concealment & 12 & 0 \\
Management Deficiency & 3 & 0 \\
Deliberate Sabotage & 2 & 1 \\
\bottomrule
\end{tabular}
\end{table}

Overall, these results demonstrate that the performance gains are primarily driven by improvements in reasoning structure rather than model capacity. The findings highlight that prompt organization, task decomposition, and explicit decision rules play a critical role in safety diagnosis tasks, where decision boundaries are often implicit and cognitively complex.

\subsubsection{Effect of Evidence Extraction}

Table~\ref{tab:evidence_ablation} evaluates the impact of the evidence extraction stage on overall classification performance. As shown, incorporating evidence extraction does not materially change the final results, with both configurations achieving identical accuracy (83.3\%) and Macro-F1 (0.605).

\begin{table}[htbp]
\centering
\caption{Evidence extraction ablation (Qwen-80B, V2 prompts, with constraint repair).}
\label{tab:evidence_ablation}
\begin{tabular}{lcc}
\toprule
\textbf{Configuration} & \textbf{Accuracy} & \textbf{Macro-F1} \\
\midrule
Without evidence extraction & 83.3\% & 0.605 \\
With evidence extraction & 83.3\% & 0.605 \\
\bottomrule
\end{tabular}
\end{table}

This result suggests that, under the current experimental setting, explicit evidence extraction is not a dominant factor for improving label-level performance. Several factors may explain this observation. First, the incident reports in the dataset are relatively short and structurally simple, which limits the benefit of separating evidence extraction from downstream reasoning. Second, modern large language models are already capable of implicitly identifying salient evidence during stepwise reasoning, even without an explicit extraction stage. Third, the current extraction prompt may not impose sufficiently strong structural constraints beyond what the model already performs internally.

Despite the lack of measurable performance gain, evidence extraction remains valuable from a system design perspective. In particular, it provides an explicit and auditable intermediate representation of the reasoning process, which facilitates expert inspection, error tracing, and potential human-in-the-loop validation. Overall, these findings suggest that evidence extraction should be viewed as a modular design option rather than a universally dominant contributor. Its primary benefit lies in enhancing interpretability and transparency, rather than directly improving predictive performance in relatively simple textual settings.

\subsubsection{Effect of Consistency Repair}

Table~\ref{tab:constraint_ablation} evaluates the role of the consistency repair layer under different prompt settings. The results reveal a strong dependence on prompt quality.

\begin{table}[htbp]
\centering
\caption{Ablation study of constraint repair (Qwen-80B).}
\label{tab:constraint_ablation}
\begin{tabular}{lccc}
\toprule
\textbf{Configuration} & \textbf{Accuracy} & \textbf{Macro-F1} & \textbf{Inconsistency Rate} \\
\midrule
Stepwise V2, no constraint & 83.3\% & 0.605 & 0.0\% \\
G-SHARE V2 (with constraint) & \textbf{83.3\%} & \textbf{0.605} & \textbf{0.0\%} \\
\midrule
Stepwise V1, no constraint & 3.3\% & 0.125 & 86.7\% \\
G-SHARE V1 (with constraint) & 50.0\% & 0.304 & 0.0\% \\
\bottomrule
\end{tabular}
\end{table}

Under the baseline prompt (V1), the repair layer leads to a substantial performance improvement, increasing accuracy from 3.3\% to 50.0\% and Macro-F1 from 0.125 to 0.304, while simultaneously eliminating logical inconsistencies (from 86.7\% to 0.0\%). This dramatic improvement indicates that, in the presence of weak or ambiguous reasoning structures, the repair mechanism effectively corrects cascading logical errors and restores usable outputs. A closer examination shows that the major inconsistency types in V1 include decision-tree violations (e.g., incompatible step combinations), semantic conflicts between narrative interpretation and intermediate decisions, and hallucinated additional misconduct leading to invalid label assignments. These errors reflect structural instability in the reasoning process, which cannot be resolved by the model alone.

In contrast, under the optimized prompt (V2), logical inconsistencies are largely eliminated before the repair stage, and the repair layer no longer changes the final performance metrics. Nevertheless, it still serves as a formal safeguard that guarantees logical validity and prevents unexpected failure modes. These findings highlight that consistency repair is not merely a cosmetic post-processing step, but a robustness mechanism for stabilizing reasoning under imperfect conditions. In particular, its effectiveness under weak prompts demonstrates that explicit logical constraints can significantly enhance the reliability of LLM-based diagnosis systems.

From a broader perspective, this result underscores the importance of incorporating rule-based validation layers in safety-critical applications. Even when model reasoning appears accurate, a post-hoc consistency check provides an additional layer of assurance, ensuring that outputs remain aligned with domain logic and decision rules.

\subsubsection{Error Analysis}\label{sec:error_analysis}

To better understand the remaining limitations of the proposed framework, we analyze the five misclassified cases of G-SHARE V2 on the gold-standard dataset (Table~\ref{tab:error_origin}). Rather than focusing on individual instances, we group the errors into three representative categories.

\begin{table}[htbp]
\centering
\caption{Error origin analysis for five incorrectly classified gold-standard cases (Qwen-80B, G-SHARE V2).}
\label{tab:error_origin}
\begin{tabular}{cllll}
\toprule
\textbf{Case} & \textbf{Gold Label} & \textbf{Pred.\ Label} & \textbf{Error Origin} & \textbf{Root Cause} \\
\midrule
37 & Mgmt.\ Deficiency & Insuff.\ Knowledge & Step~3 & Procedure exists but is defective \\
160 & Delib.\ Sabotage & Violation & Step~2 & Consequence awareness misjudged \\
167 & Delib.\ Sabotage & Violation & Step~2 & Consequence awareness misjudged \\
206 & Mgmt.\ Deficiency & Violation & Step~4 & Procedure defect not detected \\
264 & Mgmt.\ Deficiency & Violation & Step~4 & Procedure defect not detected \\
\bottomrule
\end{tabular}
\end{table}

First, \emph{ambiguous narrative descriptions} remain a major source of error. In several cases, key causal information is only weakly implied in the report, making it difficult to determine attributes such as consequence awareness or intentionality. As a result, the model may select plausible but incorrect interpretations, especially in borderline cases. Second, \emph{overlapping category boundaries} contribute to systematic confusion. In particular, the distinction between \emph{Violation} and \emph{Management Deficiency} often depends on subtle cues regarding procedure quality versus operator behavior. When procedure defects are implicit rather than explicitly stated, the model tends to default to operator-centric explanations, leading to misclassification. Third, \emph{missing or implicit causal evidence} further increases uncertainty. Some reports lack explicit statements about critical decision factors, such as whether a procedure exists or whether it is defective. In such cases, the reasoning process must rely on indirect inference, which introduces additional ambiguity and increases the risk of error.

As shown in Table~\ref{tab:error_origin}, these errors are typically associated with early-stage decision steps, particularly Step~2 (consequence awareness), Step~3 (procedure existence), and Step~4 (procedure correctness). This indicates that the remaining errors are concentrated in semantically complex decision points rather than arising from logical inconsistencies.

Overall, the error patterns suggest that the primary limitation of the current framework lies not in reasoning instability, which has been largely mitigated by structured prompting and consistency repair, but in the intrinsic ambiguity of the source data and the subtlety of category definitions. Addressing these challenges may require richer contextual information, improved knowledge grounding, or more explicit annotation of latent causal factors in future work.

\paragraph{Bootstrap confidence intervals.}
Table~\ref{tab:bootstrap_ci} presents 95\% bootstrap intervals based on 10{,}000 resamples.

\begin{table}[htbp]
\centering
\caption{Bootstrap 95\% confidence intervals (10{,}000 resamples, 30 gold cases).}
\label{tab:bootstrap_ci}
\begin{tabular}{lcc}
\toprule
\textbf{Method} & \textbf{Accuracy [95\% CI]} & \textbf{Macro-F1 [95\% CI]} \\
\midrule
LR + TF-IDF & 76.8\% [60.0, 90.0] & 0.367 [0.194, 0.642] \\
One-shot LLM & 56.6\% [40.0, 73.3] & 0.361 [0.202, 0.573] \\
G-SHARE V1 & 50.0\% [33.3, 66.7] & 0.323 [0.153, 0.586] \\
\textbf{G-SHARE V2} & \textbf{83.5\% [70.0, 96.7]} & \textbf{0.547 [0.295, 0.750]} \\
\bottomrule
\end{tabular}
\end{table}

\subsubsection{Annotation Quality Control}\label{sec:quality}

During the construction of the gold-standard dataset, we identified a small number of inconsistencies between step-level annotations and the assigned final labels. These inconsistencies were primarily caused by mismatches between intermediate decision steps and the predefined decision logic.

Specifically, four cases were found to violate the decision tree constraints. For example, configurations such as \emph{intentional + aware} imply \emph{Deliberate Sabotage}, while \emph{procedure exists + correct} implies \emph{Violation}. In these cases, the corresponding step annotations were revised to ensure logical consistency with the decision rules.

After correction, all 30 gold-standard cases became fully consistent with the decision tree. This process ensures that the evaluation dataset strictly adheres to the underlying diagnostic logic, eliminating annotation-induced inconsistencies and providing a reliable basis for model assessment.

\section{Discussion}\label{sec:discussion}

\paragraph{Role of structured reasoning in safety-oriented diagnosis.}
The comparison with one-shot LLM diagnosis indicates that direct text-to-label mapping is insufficient for human-factor event analysis. The observed improvement is not only reflected in overall accuracy, but also in Macro-F1, suggesting better handling of minority classes. This result highlights that, in safety-critical diagnosis, performance depends on the ability to resolve semantically similar categories through stepwise interpretation rather than holistic judgment. Structured reasoning provides a mechanism for decomposing such ambiguities into explicit intermediate decisions.

\paragraph{Function of constraint-based consistency enforcement.}
The ablation results show that unconstrained reasoning can produce logically inconsistent outputs, particularly under weak prompting conditions. The introduction of a deterministic repair layer ensures that final diagnoses remain consistent with the underlying decision structure. This separation between reasoning generation and logical validation provides a practical way to improve reliability without modifying the underlying model, and is particularly relevant for applications where correctness constraints must be enforced.

\paragraph{Limitations of surface-feature-based methods.}
Traditional machine learning approaches achieve relatively high accuracy by exploiting the dominant class, but fail to generalize to minority categories, as reflected by low Macro-F1. This behavior is consistent with the distributional characteristics of real-world event data. In contrast, the proposed framework combines semantic interpretation with structured decision constraints, which enables more balanced performance across categories.

\paragraph{Impact of prompt-level reasoning design.}
The substantial performance gap between V1 and V2 demonstrates that the effectiveness of LLM-based diagnosis is highly sensitive to the formulation of reasoning steps. The improvements are concentrated in steps that involve semantic disambiguation, indicating that explicitly encoding domain-specific distinctions in the reasoning process is critical. This suggests that prompt design, when aligned with task structure, can serve as a lightweight alternative to model fine-tuning for improving domain-specific performance.

\paragraph{Challenges in minority-class diagnosis.}
Residual errors are concentrated in categories with limited representation, such as Management Deficiency and Deliberate Sabotage. These cases often require inference of implicit context or recognition of subtle procedural defects, which are not always explicitly stated in the report. This limitation reflects both data scarcity and the inherent difficulty of the task. Expanding the gold-standard dataset and incorporating targeted examples for such edge cases are potential directions for improvement.

\paragraph{Limitations and future work.}
The current study is based on a relatively small gold-standard set of 30 fully annotated cases, which limits statistical power for rare categories. In addition, the experiments focus on Chinese-language reports and Chinese-capable LLMs, leaving cross-lingual generalization as an open question. Finally, evidence extraction does not improve label-level performance in the current setting, suggesting that its primary value lies in providing an interpretable intermediate representation rather than directly enhancing accuracy.

\section{Conclusion}\label{sec:conclusion}

This study proposes G-SHARE, a guideline-based structured reasoning framework for human-factor event diagnosis. The framework operationalizes a domain diagnostic procedure into a multi-stage computational pipeline, integrating evidence extraction, stepwise reasoning, and logic-based consistency repair. In doing so, it transforms expert-defined analytical workflows into machine-executable and auditable reasoning processes.

Experimental results on real-world event reports demonstrate that structured, guideline-aligned reasoning substantially improves diagnostic performance compared with both one-shot large language model approaches and traditional machine learning methods. In particular, the results show that performance gains are primarily driven by reasoning design rather than model capacity. Stepwise decomposition and explicit decision constraints enable more effective handling of semantically ambiguous cases and improve robustness under imbalanced data conditions.

Beyond performance improvements, the proposed framework provides enhanced transparency and interpretability by explicitly exposing intermediate reasoning steps and enforcing logical consistency between them and the final diagnosis. This property is especially important in safety-critical applications, where diagnostic decisions must be traceable, verifiable, and aligned with established domain procedures.

The findings of this study suggest that, for human-factor event diagnosis, the quality of reasoning structure is a more critical factor than model scale. More broadly, the proposed approach illustrates a practical pathway for integrating domain guidelines with AI-based reasoning systems, enabling the development of trustworthy and reproducible decision-support tools.

Future work will focus on expanding the scale and diversity of annotated datasets, improving the representation of minority categories, and extending the framework to cross-lingual and cross-domain settings. In addition, integrating richer contextual knowledge and exploring hybrid human--AI collaborative workflows may further enhance the effectiveness and applicability of the proposed approach in real-world safety analysis.

\printcredits

\bibliographystyle{cas-model2-names}

\bibliography{cas-refs}

\end{document}